\documentclass[lettersize,journal]{IEEEtran}
\usepackage{amsmath,amsfonts,amssymb}
\usepackage[ruled]{algorithm2e}
\usepackage{array}
\usepackage{bbm}
\usepackage[caption=false,font=normalsize,labelfont=sf,textfont=sf]{subfig}
\usepackage{textcomp}
\usepackage{stfloats}
\usepackage{url}
\usepackage{verbatim}
\usepackage{graphicx}
\usepackage{booktabs}
\usepackage{multirow} 
\usepackage{cite}
\usepackage{hyperref} 

\makeatletter

\newcommand{\Rmnum}[1]{\expandafter\@slowromancap\romannumeral #1@}
\makeatother

\begin{document}

\title{

Towards Socially Responsive Autonomous Vehicles: A Reinforcement Learning Framework with Driving Priors and Coordination Awareness
}

\author{Jiaqi Liu,~\IEEEmembership{Student Member,~IEEE,} Donghao Zhou, Peng Hang,~\IEEEmembership{Member,~IEEE,} Ying Ni, and Jian Sun
\thanks{This work was jointly supported by the National Natural Science Foundation of China (52302502, 52125208, 52272313), the National Key R\&D Program of China (2022YFB2502901), the Young Elite Scientists Sponsorship Program by CAST (2022QNRC001) and the Fundamental Research Fund for the Central Universities.}
\thanks{Jiaqi Liu, Donghao Zhou, Peng Hang, Ying Ni, and Jian Sun are with the Department of
Traffic Engineering and Key Laboratory of Road and Traffic Engineering,
Ministry of Education, Tongji University, Shanghai 201804, China. (e-mail: \{liujiaqi13, zhoudonghao, hangpeng, ying\_ni, sunjian\}@tongji.edu.cn)}

\thanks{Corresponding author: Peng Hang}
}

\markboth{}%
{Shell \MakeLowercase{\textit{et al.}}: A Sample Article Using IEEEtran.cls for IEEE Journals}

\maketitle 

\begin{abstract}
The advent of autonomous vehicles (AVs) alongside human-driven vehicles (HVs) has ushered in an era of mixed traffic flow, presenting a significant challenge: the intricate interaction between these entities within complex driving environments. AVs are expected to have human-like driving behavior to seamlessly integrate into human-dominated traffic systems. 
To address this issue, we propose a reinforcement learning framework that considers driving priors and Social Coordination Awareness (SCA) to optimize the behavior of AVs. The framework integrates a driving prior learning (DPL) model based on a variational autoencoder to infer the driver's driving priors from human drivers' trajectories. A policy network based on a multi-head attention mechanism is designed to effectively capture the interactive dependencies between AVs and other traffic participants to improve decision-making quality. The introduction of SCA into the autonomous driving decision-making system, and the use of Coordination Tendency (CT) to quantify the willingness of AVs to coordinate the traffic system is explored. Simulation results show that the proposed framework can not only improve the decision-making quality of AVs but also motivate them to produce social behaviors, with potential benefits for the safety and traffic efficiency of the entire transportation system. 
\end{abstract}

\begin{IEEEkeywords}
Autonomous Vehicle; Reinforcement Learning;  Driving Prior Learning; Autoencoder; Social Coordination
\end{IEEEkeywords}

\section{Introduction}
The proliferation of autonomous vehicles (AVs) into real-world road traffic systems, coexisting alongside human-driven vehicles (HVs), is ushering in a new era of human-machine mixed traffic flow\cite{wang2023new,negash2022anticipation,hang2022conflict}. This transformative shift, while promising, introduces a formidable challenge: the intricate interaction between AVs and HVs within these complex driving environments\cite{chen2022milestones}. The confluence of AV-HV dynamics poses fresh demands on AVs, demanding heightened interaction capabilities and adaptability in the face of environmental intricacies.

In many intricate interaction scenarios, AVs often struggle to emulate the nuanced social interactions innate to human drivers during their commute\cite{hang2023brain}. To seamlessly integrate into traffic systems governed by human participation, AVs must exhibit a keen acumen for discerning real-time human driver intentions and adeptly manifest social behaviors akin to their human counterparts~\cite{hang2022decision}. This encompasses emulating the actions of fellow traffic participants, assimilating an array of human driving traits, and engaging in a manner comprehensible to human drivers. We term this phenomenon "social coordination".

Social coordination is a ubiquitous facet of human behavior, surfacing prominently in social interactions and team-based activities. The propensity for humans to harmoniously coordinate with one another to achieve common objectives has long been a focal point in kinematic research~\cite{VILAR201414}. The pursuit of enabling autonomous vehicles to proficiently coordinate with humans has garnered considerable attention from researchers~\cite{wang2021socially,schwarting2019social,lu2022autonomous}.

Reinforcement Learning (RL), renowned for its efficiency and scalability, has emerged as a pivotal tool in the training of decision-making algorithms for autonomous driving~\cite{he2023robust,saxena2020driving,liu2021reinforcement}. Nevertheless, the reliance on reward functions and opaque policy networks has presented challenges in areas such as intention comprehension, anthropomorphism, and interpretability.

To address these challenges, we proffer a RL framework that synergistically incorporates driving priors and social coordination awareness (SCA) to orchestrate AVs' navigation, engendering actions that are efficient, safe, socially adept, and imbued with anthropomorphic attributes.
Our framework is underpinned by a driving prior learning (DPL) model, harnessing the capabilities of a variational autoencoder to glean driving prior information from HVs' trajectories. Augmenting this architecture, we infuse an attention mechanism into the policy network, fashioning a multi-head attention-based policy network. This design imbues the network with the ability to proficiently apprehend interactive dependencies intrinsic to the interplay between AVs and other traffic participants, elevating the caliber and dependability of decision-making. Notably, the Proximal Policy Optimization (PPO) algorithm~\cite{schulman2017proximal} undertakes the mantle of learning and iteratively enhancing the policy network.

Moreover, we introduce the novel concept of SCA into the fabric of autonomous driving decision-making. SCA is operationalized through the coordination tendency (CT), a quantitative metric gauging AVs' inclination toward engaging in traffic system coordination. This endeavor is complemented by the design of a bespoke reward function and a comprehensive exploration of AV performance across varying CTs, meticulously analyzing their impact on the broader traffic ecosystem. Our empirical findings underscore the potential of our framework to not only enhance the decision-making prowess of AVs but also to stimulate the emergence of socially conscious behaviors. This, in turn, holds the promise of elevating the safety and traffic efficiency of the entire transportation landscape.

Our contributions are summarized as follows:
\begin{itemize}
     \item A RL decision-making framework that considers driving priors and social coordination awareness is proposed for AVs, advancing the social interaction abilities of AVs;
     \item A Variational Autoencoder (VAE)-based DPL model is used  to predict the human drivers' styles and intentions, which are sent to the policy network as prior information to assist in action decision-making, and a policy network based on the multi-head attention mechanism is designed to capture the interaction dependencies between AVs and other traffic participants;
     \item SCA is introduced to encourage the generation of socially responsible behaviors in AVs, and the coordination tendency is utilized to measure the willingness of autonomous vehicles to coordinate the traffic system.
\end{itemize}

The rest of the paper is organized as follows: Section~\ref{section:2} summarizes the recent related works. The decision-making problem and our whole framework are described in section~\ref{section:4}. Section~\ref{section:4} describes some preliminaries of our work.  In section~\ref{section:5}, the framework we proposed is described. In section \ref{section:6}, the simulation environment and comprehensive experiments are introduced and the results are analyzed. Finally, this paper is concluded in section \ref{section:7}.

\section{Related Works}
\label{section:2}
\subsection{Decision-Making of AVs}
An intelligent decision-making system is critical for the safe and efficient driving of AVs. Decision-making strategies and algorithms for AVs have been widely studied in recent years, including rule-based methods\cite{zhang2017finite}, game theory-based methods\cite{cai2021game}, and learning-based methods\cite{peng2021end,liu2023cooperative}. 

Demonstrating robust learning capabilities coupled with high-fidelity inference execution, RL has garnered substantial traction in the conceptualization and training of decision-making algorithms, a trend accentuated by the works of Jin et al. and Shu et al.~\cite{jin2020game,shu2021driving}. Saxena et al.~\cite{saxena2020driving} notably introduced a model-free RL approach that facilitates the acquisition of a continuous control policy across the action space of AVs, thereby effectuating a discernible enhancement in the safety quotient of AVs operating within dense traffic environs.
In the context of intricate intersection scenarios, widely acknowledged as among the most intricate driving conditions, Liu et al.~\cite{liu2021reinforcement} conceived a comprehensive RL benchmark, meticulously tailored to cater to the exigencies of AVs' decision-making within such complex contexts. Furthermore, the versatility of the RL paradigm is underscored by its deployment as a foundational framework for training multi-task decision-making models specific to AVs, as illustrated by Liu et al.~\cite{liu2023mtd}.

However, it is pertinent to note that RL, hinging on extensive exploration and iterative trial-and-error, is confronted with inherent challenges related to learning efficiency. To address this, the present study seeks to augment the learning efficiency of AVs through the assimilation of prior driving reasoning and attention mechanisms.

\subsection{Driving Latent States Modeling and Inference}
The modeling and inference of latent driving states within HVs, encompassing driving styles and driving intentions, holds paramount significance for the advancement of AVs. These endeavors are pivotal for enhancing AVs' capacity to comprehend and anticipate the behavioral patterns exhibited by HVs. Diverse methodologies have been deployed to effectively model latent driving states, encompassing both unsupervised and supervised learning techniques. Noteworthy among these are approaches such as the hidden Markov model (HMM) \cite{song2016intention}, graph neural networks (GNN) \cite{ma2021reinforcement}, and inverse reinforcement learning (IRL) \cite{kuderer2015learning}. It is imperative to note that the nature of drivers' styles and intentions often remains implicit and devoid of direct communication or observation. Concurrently, the endeavor to amass an ample volume of driving state labels from HVs in real-world settings poses considerable challenges to supervised learning paradigms. As a result, the application of unsupervised learning methodologies, notably variational inference \cite{morton2017simultaneous,schmerling2018multimodal}, has emerged as a highly promising avenue.

In the present study, we employ an unsupervised model founded on VAEs, termed the DPL model, to adeptly apprehend and deduce the latent driving states of HVs. 

\subsection{Social Behavior in Autonomous Driving}
During the driving process, there will be interactive behaviors between human drivers including cooperation, gaming, and polite yielding. It is necessary for AVs to learn these behaviors in order to behave more like humans, which is crucial for AVs to integrate into the human world.

In recent years, some works have focused on extracting the interaction information and relationship of traffic participants, such as Social LSTM \cite{alahi2016social}, GNN \cite{li2021interactive}, to improve the decision-making quality of AVs.  
Some works quantify and estimate the degree of the sociality of AVs and HVs\cite{schwarting2019social,wang2021socially}.
Hang et al.~\cite{hang2022decision} proposed a game-theoretic decision-making framework for the unsignalized intersection scenarios that can advance social benefits, including the traffic system efficiency and safety, as well as the benefits of AVs.

Schwarting et al. \cite{schwarting2019social} pioneered the utilization of Social Value Orientation (SVO) to quantify the spectrum of an agent's self-interest and altruism, thereby influencing the decision-making demeanor of AVs. 
Wang et al. \cite{wang2021socially} devised an online prediction algorithm to infer the behavioral traits of fellow traffic participants, effectively guiding AVs in the generation of judicious and psychologically-congruent decision-making actions. 
Behrad et al. \cite{toghi2022social} interwove SVO into the Multi-Agent Reinforcement Learning (MARL) framework, probing the coalescing effect of Connected and Autonomous Vehicles (CAVs) on the transportation landscape within the ramp-in scenario.
Zhao et al. \cite{zhao2021yield} introduced an active semantic decision-making methodology, predicated on a game theory model incorporating quantifiable social preferences and counterfactual reasoning. 
Crosato et al. \cite{crosato2021human} artfully integrated SVO into the RL paradigm, tailoring AV behaviors towards pedestrians from audacious to prudent. 

In alignment with these scholarly endeavors, our work introduces the concept of SCA, wherein the AV's inclination towards coordination is modulated by a coordination tendency. 

\section{Problem Statement and Framework Overview}
\label{section:3}
\subsection{Scenario Description}

We delve into the intricacies of decision-making pertaining to a solitary AV operating within an unsignalized intersection scenario. 
The absence of traffic signal control in such intersections introduces ambiguity regarding the right of way, imposing heightened demands on vehicular interaction capabilities~\cite{hang2022decision}. Consequently, this scenario serves as a rigorous testbed for evaluating the social performance of AVs.
The focal point of our analysis revolves around a single-lane cross-shaped unsignalized intersection, a locale characterized by the potential presence of multiple HVs. These HVs may emanate from disparate directions and origins, each imbued with its unique driving styles and intentions.

\begin{figure}[!htbp]
    \centering
    \includegraphics[width=0.4\textwidth]{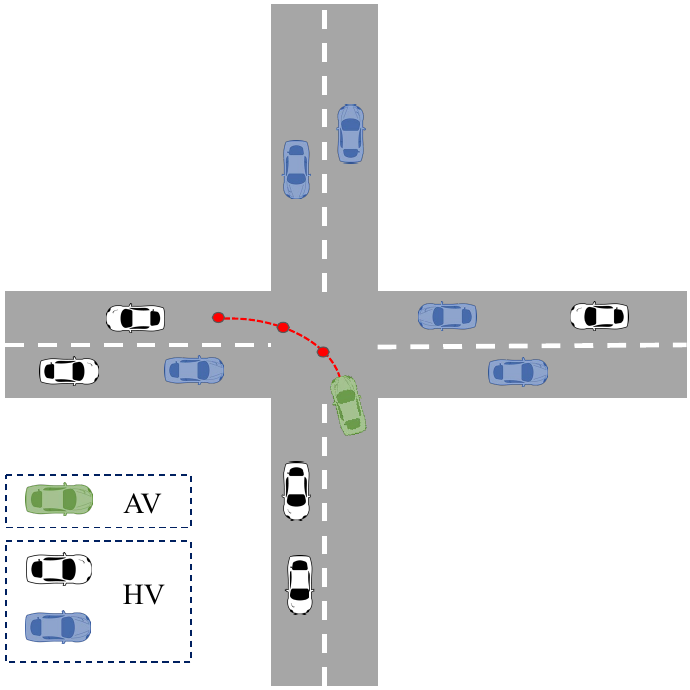}
    \caption{The unsignalized intersection scenario.}
    \label{fig:Origin_Scenario}
\end{figure}

The principal objective guiding our AV is to safely traverse the intersection and reach its designated destination, a mission it must accomplish adeptly. The scenario under scrutiny is graphically represented in Fig.~\ref{fig:Origin_Scenario}.

The high-level actions of the AV in our problem are dictated by the RL algorithm, which subsequently translates these decisions into precise low-level steering and acceleration signals utilizing a closed-loop Proportional-Integral-Derivative (PID) controller~\cite{bacha2017review}. The motion of the vehicle is delineated through the application of a Kinematic Bicycle Model~\cite{polack2017kinematic}.

\begin{figure*}[!htbp]
    \centering
    \includegraphics[width=0.9\textwidth]{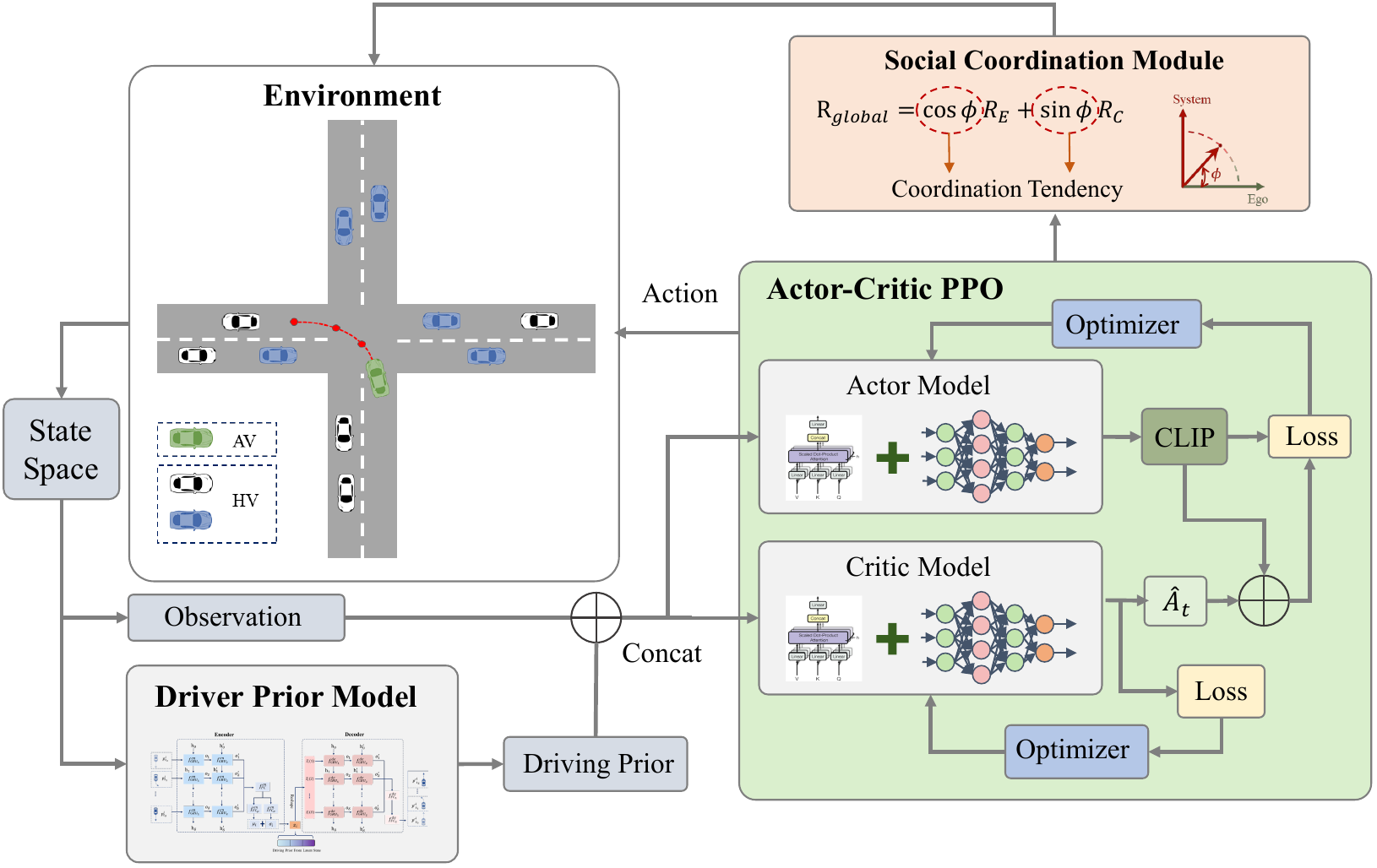}
    \caption{The RL-based decision-making framework for socially responsive autonomous driving system.
    }
    \label{fig:PPO_model_overview}
\end{figure*}

\subsection{Framework Overview}
The overarching decision-making framework we propose, as depicted in Fig.~\ref{fig:PPO_model_overview}, operates within the realm of RL and is meticulously engineered to encompass driving priors and SCA. This framework serves as the lodestar guiding AVs toward the generation of pro-social and anthropomorphic interactive behaviors.

At its core, the framework seamlessly integrates a DPL model, hinging on a VAE, adept at inferring driving prior information from human-driven vehicle trajectories. Concomitantly, a multi-faceted policy network, underpinned by the multi-head attention mechanism, assumes the pivotal role of action policy learning. The PPO algorithm~\cite{schulman2017proximal} is adroitly employed for policy network learning and continuous updates.

Additionally, we introduce SCA into the autonomous driving decision-making paradigm. To quantitatively encapsulate the AVs' inclination toward participating in traffic system coordination, we harness the CT metric. This holistic framework orchestrates a symphony of predictive learning, interactive policy optimization, and cooperative orientation, charting a course towards the realization of socially aware and harmoniously interacting AVs.

\section{Preliminaries}
\label{section:4}

\subsection{Partially Observable Markov Decision Process}
The sequential decision-making process of an AV in the dynamic environment can be described by a partially observable Markov Decision Process (POMDP)~\cite{spaan2012partially}. POMDP can be specified by the tuple $(\mathcal{S}, \Omega, \mathcal{A}, \mathcal{T}, \mathcal{R}, \gamma, \rho_0)$, where $\mathcal{S}$ is the state space; $\mathcal{A}$ is the action space; $\mathcal{T}$: $\mathcal{S} \times \mathcal{A} \times \mathcal{S} \rightarrow \mathbb{R}$ is the transition model; $\mathcal{R}: \mathcal{S} \times \mathcal{A} \rightarrow \mathbb{R}$ is the reward function; $\gamma \in [0,1]$ is the discount factor; and $\rho_0: \mathcal{S} \rightarrow \mathbb{R}$ is the initial state distribution.
$\Omega$ is used to map a state $s\in \mathcal{S}$ to an observation $o\in \mathcal{O}$, where $\mathcal{O}$ is the observation space.

In the POMDP, the agent makes decisions and takes actions according to the policy $\pi$ and the current observation. The goal of the agent is to find the optimal policy $\pi^*$ maximizing expected $\gamma$- discounted cumulative reward.  Formally, the value function $V_{\phi}(s)$ and the action value function $Q_{\phi}(s,a)$ are defined as:
\begin{equation}
    \begin{aligned}
        & V^{\pi} (s) \overset{\text{def}}{=} \mathbb{E}[\sum_{t=0}^\infty \gamma^t R(s_t,a_t ) \lvert s_0=s,a_t \sim \pi(a_t \lvert s_t ),\\
        & s_{t+1} \sim P(s_{t+1},a_t) ]
    \end{aligned}
\end{equation}
\begin{equation}
    Q^{\pi} (s,a) \overset{\text{def}}{=}
    R(s,a)+\gamma E_{s^\prime \sim P(s^\prime \lvert s,a) } V^\pi (s^\prime)
\end{equation}

The optimal action-value function $Q^{*}=\max_{\pi} Q^{\pi}(s)$ satisfies the Bellman Optimality Equation:
\begin{equation}
    Q^* (s,a)  \overset{\text{def}}{=} E_{s^\prime \sim P(s^\prime \lvert s,a) } \max_{a^\prime \in A}[R(s,a)+\gamma Q^* (s^\prime,a^\prime)]
\end{equation}

\subsection{Proximal Policy Optimization}
PPO~\cite{schulman2017proximal}, as a policy gradient method, performs very well in many challenging environments. 

The main idea of PPO-Clip is the clipping surrogate objective:
\begin{equation}
    \begin{aligned}
        L^{\text{PPO}}(\theta)=E_t \Big[ \min \Big( r_t(\theta)\hat{A_t}, &\\
        \text{clip} \big (r_t(\theta),1-\epsilon,1+\epsilon \big) \hat{A_t} \Big) \Big]
    \end{aligned}
    \end{equation}
where $r(\theta) = \frac{\pi_{\theta}(a|s)}{\pi_{\theta ^{\prime}(a|s)}}$ denotes the ratio of the new policy $\pi_{\theta}(a|s)$ to the old policy $\pi_{\theta ^{\prime}(a|s)}$,  $\hat{A_t}$ signifies the advantage function and $\epsilon$ is the clipping range.

\subsection{Autoencoder}
An Autoencoder (AE) is a neural network that is trained in an unsupervised way to reproduce a model's input to its output, whose goal is to minimize the reconstruction error~\cite{bank2023autoencoders}. An AE usually contains two main components: an encoder and a decoder. The encoder maps the input $x$ to a latent feature representation $z$, which is denoted by $z = g_{\phi}(x)$. The decoder obtains a reconstruction $y$ of the input $x$ by using the latent feature representation $z$, which is denoted by $y = f_{\theta}(z)$. The difference between output $y$ and input $x$ is used as the objective function. For example, Mean Squared Error (MSE) is frequently used as the loss function:
\begin{equation}
    \mathcal{L}_{AE}(x) = || x - f_{\theta}(g_{\phi}(x)) || ^ 2  = || x - f_{\theta}(z) || ^ 2
\end{equation}

\section{Methodology}
\label{section:5}
This section details how we improve the social interaction ability of autonomous vehicles. First, the observation space and action space of our algorithm is described in detail. Then we introduce the DPL model and social coordination module.

\begin{figure*}
    \centering
    \includegraphics[width=0.9\textwidth]{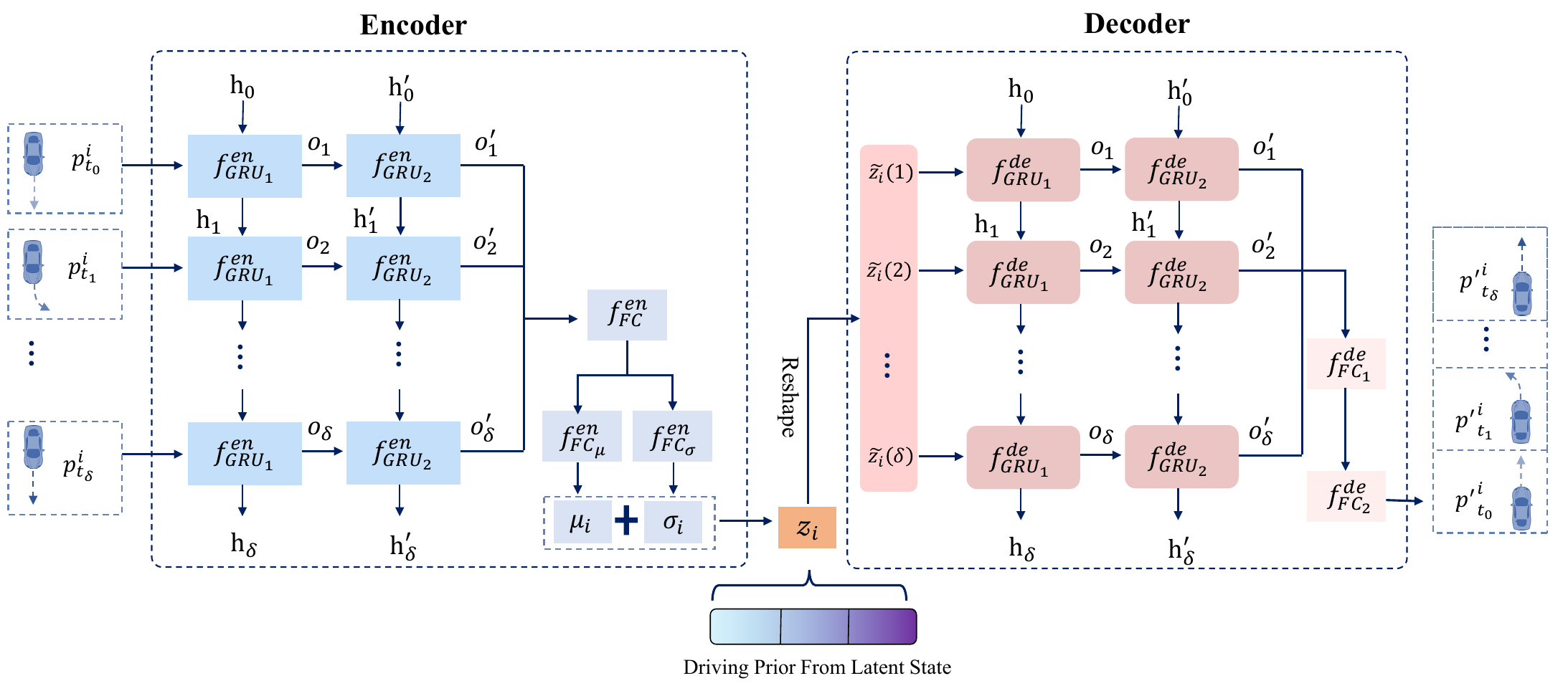}
    \caption{The network architecture of the DPL model.
    }
    \label{fig:VAE_model}
\end{figure*}

\subsection{Observation and Action Space}
\subsubsection{Observation Space}

Let $\mathcal{N}_i$ denote the set of all observable vehicles within the perceptual scope of agent $i$. The observation matrix of agent $i$, denoted as $\mathcal{O}_i$, exhibits dimensions $|\mathcal{N}_i| \times |\mathcal{F}|$, where $|\mathcal{N}_i|$ represents the count of observable vehicles for agent $i$, and $|\mathcal{F}|$ signifies the number of features employed to encapsulate a vehicle's state. The feature vector for vehicle $k$ can be succinctly expressed as:
\begin{equation}
\mathcal{F}_k = [x_k, y_k, v^x_k, v^y_k]
\end{equation}
where $x_k$, $y_k$, $v^x_k$, and $v^y_k$ correspond to the longitudinal position, lateral position, longitudinal velocity, and lateral velocity, respectively.

\subsubsection{Action Space}
This research delves into the exploration of social coordination dynamics and interactions between AVs and HVs. Consequently, the focus lies predominantly on the high-level decision-making actions of AVs, rather than delving into the minutiae of vehicle-level control. Within the context of traversing intersections, where predetermined routes are in place, AVs are tasked with determining acceleration and deceleration actions required to execute left turns and reach their destinations. The action space $\mathcal{A}$ for AVs is thus defined as the encompassing set of high-level control decisions, encompassing $\{ slow \ down, cruising, speed \ up \}$. Subsequent to the selection of a high-level decision, subordinated controllers effectuate the generation of corresponding steering and throttle control signals to govern the motion of AVs.

\subsection{Driving Prior Learning}
\label{section:RPL}

In complex interaction scenarios, understanding the driver's interaction intention and driving style is very helpful for AV to understand the behavior of human drivers, and can effectively improve the safety and efficiency of AV. However, in reality, the heterogeneity of drivers exists widely and their styles are vague and implicit. of. To address the above issues, we propose a DPL model to infer the human driving latent states in interactive scenarios.

 Let $p^i_{t} = [x^i_{t},y^i_{t}]$ denote the position information of HV $i$ at timestep $t$ and let $p^{\prime i}_{t}$ denote the predicted position information by the DPL model, where $x^i_{t}$ and $y^i_{t}$ are the longitudinal position and lateral position of HV $i$, respectively.
 The action distribution of HV $i$ is modeled as $P(p^{\prime i}_{t} | p^i_{t},z^i )$, where $z^i$ represents the latent driving style of the HV $i$.
 The goal of DPL model is to learn $P(z^i | p^i_{0:\delta})$, where $p^i_{0:\delta}$ is the position information of HV $i$ up to timestep $\delta$.
 
\subsubsection{Model Architecture}
 The network architecture of the DPL model is shown in Fig.\ref{fig:VAE_model}. The DPL is a VAE, which contains two modules: an encoder $g_{\phi}$ and a decoder $f_{\theta}$. 

In the encoding module, the state $p^i_t$ for HV $i$ is first embedded by a non-linear embedding layer $f_{embed}$ and then is fed to two-layer GRU module\cite{cho2014learning}:
\begin{equation}
\begin{aligned}
    & h_{t}, o_t = f^{en}_{GRU_1}(h_{t-1}, f_{embed}(x^t)) \\
    & h^{\prime}_t, o^{\prime}_t = f^{en}_{GRU_2}(h^{\prime}_{t_1},o_t)
\end{aligned}
\end{equation}
where $o_t$ and $o^{\prime}_t$ denote the update gates of two GRU modules, respectively, and $h_t$ and $h^{\prime}_t$ denotes the hidden state of the two GRU modules at time $t$. 

The update states $o^{\prime}_t$ is first fed to a fully connected layer $f^{en}_{FC}$ to encode and then fed to FC layer $f^{en}_{FC_\mu}$ and $f^{en}_{FC_\sigma}$ to get the Gaussian parameters of the latent driving style $z$:
\begin{equation}
\begin{aligned}
    & \mu_i = f^{en}_{FC_{\mu}} (f^{en}_{FC}(o^{\prime}_0,...,o^{\prime}_{\delta})) \\
    & \sigma_i = f^{en}_{FC_{\sigma}}(f^{en}_{FC}(o^{\prime}_0,...,o^{\prime}_{\delta}))
\end{aligned}
\end{equation}

And the latent driving style parameter $z_i$ of HV $i$ is sampled from $\mathcal{N}(\mu_i, \sigma)$ : $z_i = \mu_i + \epsilon \sigma_i, \epsilon \sim \mathcal{N}(0,I)$ and then $z_i$ is reshaped to a $m \times \delta$ dimension matrix, where $m$ denotes the dimensional of the latent space.

In the decode module,
\begin{equation}
\begin{aligned}
    & h_{t}, o_t = f^{de}_{GRU_1}(h_{t-1}, \tilde{z}_i(t) ) \\
    & h^{\prime}_t, o^{\prime}_t = f^{de}_{GRU_2}(h^{\prime}_{t_1},o_t)
\end{aligned}
\end{equation}

\begin{equation}
    p^{\prime i}_{t} = f^{de}_{FC_2} (f^{de}_{FC_1} (o^{\prime}_0,...,o^{\prime}_{\delta}) )
\end{equation}

\begin{figure*}
    \centering
    \includegraphics[width=0.9\textwidth]{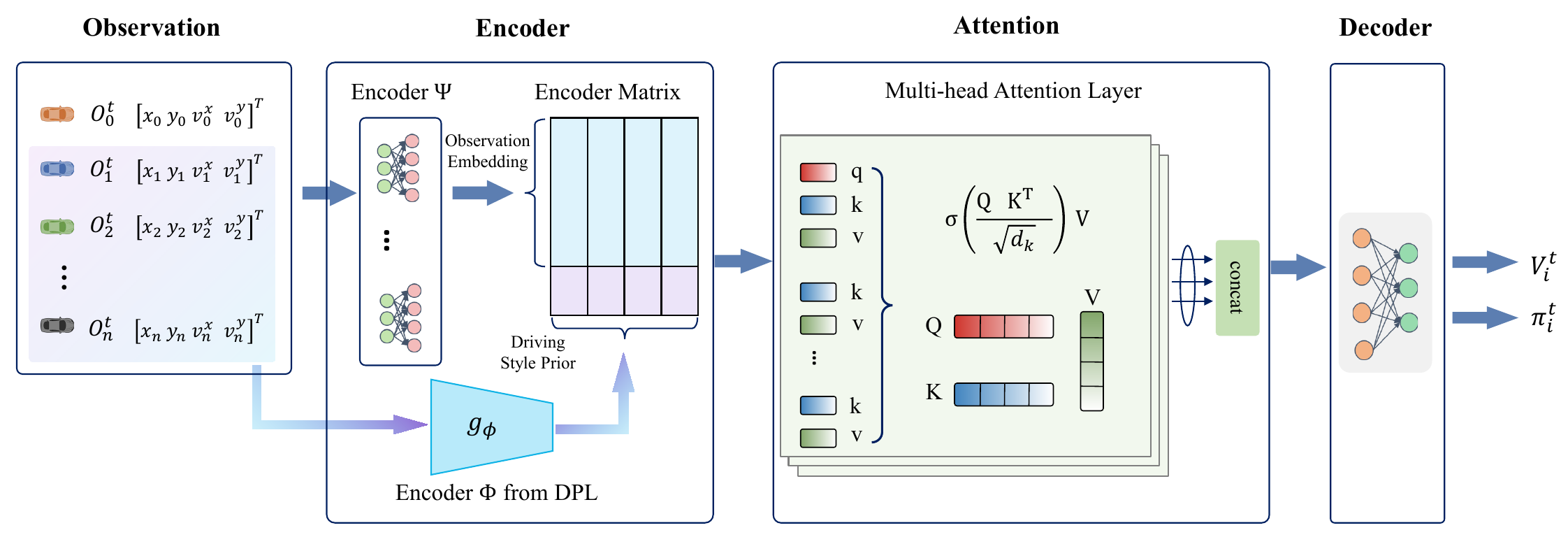}
    \caption{The architecture of our policy network for the PPO algorithm.
    }
    \label{fig:policy_network}
\end{figure*}

\subsubsection{Data Generation and Training}

To train the DPL model, the human driving data is first sampled from the simulation platform.
The human drivers with different driving styles are modeled by the IDM~\cite{kesting2010enhanced} with different parameters, which is shown in Table~\ref{tab:Driving_style_parameter}. We randomly generate drivers with different styles and different driving intentions (turn left, go straight, and turn right) at unsignalized intersections, simulate $10^4$ times and collect their trajectories. This process does not include any AVs. Then the trajectories we collect are used to train the DPL model in an unsupervised way.

\begin{table*}[htp]
\caption{The parameters of different driving styles.}
\label{tab:Driving_style_parameter}
\centering
\resizebox{\linewidth}{!}
{
\begin{tabular}{@{}ccccc@{}}
\toprule
Driving Style   & 
\begin{tabular}[c]{@{}c@{}} $\text{Jam Distance}$ \\ $(d_0)(m)$\end{tabular} & 
\begin{tabular}[c]{@{}c@{}} $\text{Desired Time}$  \\ $\text{Headway}(T)(s)$\end{tabular} &
\begin{tabular}[c]{@{}c@{}} $\text{Maximum}$ \\ $\text{Acceleration}(a_0)(m/s^2)$\end{tabular} & \begin{tabular}[c]{@{}c@{}}$\text{Maximum}$ \\$\text{Deceleration}(b_0) (m/s^2)$ \end{tabular}\\
\midrule
        Aggressive & 2.0 & 1 & 5.0 & 5.0 \\
        Moderate & 5.0 & 1.5 & 2.5 & 4.0 \\
        Conservative & 8.0 &2.0 &1.5 & 2.0 \\
\bottomrule
\end{tabular}
}
\end{table*}

In our DPL model, L2 term is used as the objective function:
\begin{equation}
    \mathcal{L}(P^i) = || P^i - f_{\theta}(z) || ^ 2 = || P^i - P^{\prime i} || ^ 2
\end{equation}
where $P^i = [p^i_{0}, p^i_{1},...,p^i_{\delta}]$ and $P^{\prime i} = [p^{\prime i}_{0}, p^{\prime i}_{1},...,p^{\prime i}_{\delta}]$.

Afterward, the inference results of the DPL model will be sent to the RL algorithm as the prior knowledge to assist the agent to make better decisions.

\subsection{Policy Network With Driving Priors and Attention Mechanism}
The structure of the policy network has a significant impact on the performance of the RL algorithm in complex environments. Due to the ability to discover the interdependencies among a variable number of inputs, the attention mechanism has been applied in the social interaction relationships extracting and analyzing. We design a policy network integrating the driving priors and attention mechanism. The overview of our policy network is shown in Fig.\ref{fig:policy_network}.

The policy network contains three modules: encoder block, attention block, and decoder block. In the encoder block, there are two encoders: encoder $\Psi$, and encoder $\Phi$. 
The features of AV $\mathcal{F}_0$ and its observation $\mathcal{O}$ are encoded by the encoder $\Psi$, which is a Multilayer Perceptron (MLP):
\begin{equation}
\label{eq:state_encoding}
    \mathcal{X} = \text{MLP}(\mathcal{F}_0,\mathcal{O})
\end{equation}

The driving prior $z_i$ of HV $i$ is inferred by:
\begin{equation}
    z_i = g_{\phi} (p^i_0,p^i_1,...,p^i_{\delta})
\end{equation}

And then the latent prior vectors from all HVs are concatenated with the feature embedding vector from encoder $\Psi$:
\begin{equation}
    \mathcal{X}^{\prime} = \text{concat}( \mathcal{X}, [ z_{1},z_{2},...,z_{n} ]^T)
\end{equation}
where $n$ is the number of HV observed by AV.

The feature matrix is fed to the attention block, which has $M$ attention heads, and the attention block produces the query results (attention weights) of AV.

In the attention block, the ego vehicle emits a single query $Q \in \mathbbm{R} ^{1\times d_k}$, where $d_k$ is the output dimension of the encoder layer. This query is then projected linearly and compared to a set of keys $K \in R^{(N+1)\times d_k} $ containing descriptive features for each vehicle. The $Q$, $K$ and $V$ are calculated as follows:
\begin{equation}
\label{eq:attention_matrix_mapping}
    \begin{aligned}
        & Q = W^Q \mathcal{X} \\
        & K = W^K \mathcal{X} \\
        & V = W^V \mathcal{X}
    \end{aligned}
\end{equation}
where the dimensions of $W^Q$ and $W^K$ are $(d_k\times d_N)$, and $W^V$'s is $(d_v \times d_h)$.

The attention weights are obtained by scaling the dot product with the inverse-square-root-dimension $\frac{1}{\sqrt{d_k}}$ and normalizing it with a softmax function $\sigma$. The attention weights are then used to gather a set of output values $V$. The attention computation for each head can be written as :
\begin{equation}
\label{eq:cal_attention_1}
    At^m = \sigma \Big (  \frac{Q K^T}{\sqrt{d_k}} \Big)V
\end{equation}

Then the output from all $M$ heads will be combined with a linear layer:
\begin{equation}
\label{eq:cal_attention_2}
    At = \sum_{m=1}^{M} At^m
\end{equation}

Finally, the vector $At$ will be fed to the decoder block, which is a MLP, to obtain the value $V^t_i$ and the policy $\pi^t_i$. Then the PPO is used to update the policy network, as shown in Fig.\ref{fig:policy_network}.

\subsection{Social Coordination Awareness of AV}
\label{sec:IO_Reward}
\begin{figure}[!htbp]
    \centering
    \includegraphics[width=0.5\textwidth]{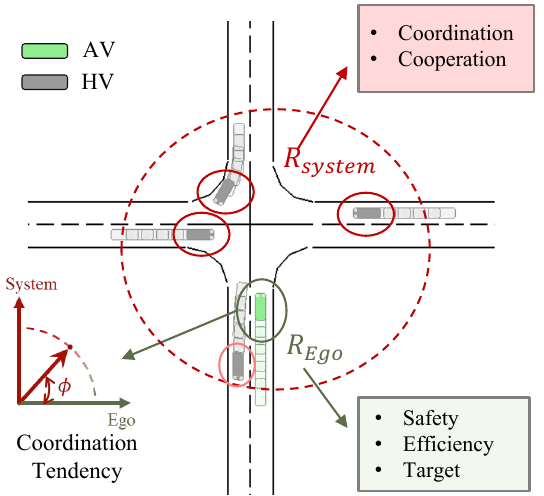}
    \caption{Social Coordination Awareness for the decision making of AV.
    }
    \label{fig:SAC_Reward}
\end{figure}

In most studies on the design of autonomous driving decision-making algorithms using reinforcement learning, the AV car only pays attention to whether its own goals are achieved, such as ensuring sufficient safety, efficiency, and comfort, and the reward function is also set and adjusted according to these goals. Such AVs are highly self-centered, ignoring the interaction with other traffic participants and the impact on the efficiency and reliability of the entire traffic system. Related studies have shown that in some scenarios, appropriate cooperation and altruism (such as slowing down and yielding) can effectively improve the overall utility of the traffic system\cite{toghi2022social}. We believe that in future human-machine mixed driving traffic scenarios, AVs are obliged to coordinate other traffic participants and improve the performance of the entire traffic system while completing their own driving goals.

In social science, social coordination involves human-machine matching with other people's thoughts, emotions, and behaviors, as well as synchronization with other people's rhythms and roles. Social coordination is a basic manifestation of human social and interactive capabilities\cite{ackerman2010two}.
We introduce SCA into autonomous driving decision-making systems.
And SCA is defined as the willingness of the AV to coordinate the behavior of both parties and the entire system based on prior understanding and the state of the interactors in the process of interacting with other traffic participants.
The global benefit $R_{global}$ is defined as the reward function for AV to promote AV's attention to the coordination of the entire system, and we used the CT) to quantify the intensity of this willingness, as shown in Fig.\ref{fig:SAC_Reward}. Similar to SVO\cite{schwarting2019social}, CT is represented by angle $\phi$, which is used to balance self-interest and coordinated system benefits.

\begin{equation}
\label{eq:R_global}
R_{global}(s,a) = \cos \phi R_{E} + \sin \phi R_{C}
\end{equation}

\begin{equation}
    R_{E} = r (s, a) = \sum_{ t\in \{ c,e,a \}} w_t*r_t 
\end{equation}

\begin{equation}
    R_{C} = \Omega (\mathcal{O})
\end{equation}
where $R_E$ is the AV's individual benefit, including safety, efficiency, goals, etc., and $R_C$ is the coordination benefit, which is estimated by the AV based on the current observation state $\mathcal{O}$ using the function $ \Omega(*)$.

The effect of AV social coordination behavior is affected by two factors: coordination propensity $\phi$, and system benefit estimation function $ \Omega(*)$.

When $\phi = 0$, AV is egoistic and won't consider the utilities of other traffic participants and the whole system; when $\phi = \frac{\pi}{2}$, AV will just maximize the reward of the whole system; in the real world, $\phi \in (0, \frac{\pi}{2} )$.

The coordination estimation function $ \Omega(*)$ is defined as:
\begin{equation}
    \Omega (\mathcal{O}) = \alpha \sum_j f(o_j)
\end{equation}
\begin{equation}
    f(o_{j}) = \frac{1}{ e ^{\lambda d_{j}}} (w_c r^j_c + w_e r^j_e)
\end{equation}
where $\alpha$ is the coefficient, $d_j$ is the distance between the AV and HV $j$. The efficiency term $r_e$ and safety term $r_c$ are considered in our simulation.
Indicators such as system conflict density and system coordination improvement will be considered in our future study.

The work process of the prior-attention PPO model is shown in Alg. \ref{alg:PA-PPO}.
\begin{algorithm}
\SetAlFnt{\small}
    \SetKwInOut{Parameter}{Inputs}
    \SetKwInOut{Output}{Output}
\caption{Prior-Attention PPO Model}
\label{alg:PA-PPO}
\LinesNumbered
\SetAlgoLined
\Parameter{Initial policy parameters $\theta_0$, initial value function parameters $\phi_i$, $\mathcal{O}$ }
\Output{$\theta$, action $a$}
\hrule
\vspace{0.2em}

\For{$Episode = 1$ to $M$}
{
{\bf Initialize} replay buffer $\mathcal{D} \leftarrow \emptyset$;\\
\For{t=0,1,$\cdots$ to Iteration Times $T_{max}$}
{
Get observation $o_{t}$\;
Obtain driving prior vector $\mathcal{Z}_t$ by DPL model $g_{\phi}$ and $o_{t}$;\\
Get action distribution $p(a)$ by running policy $\pi$ with $o_t$ and $\mathcal{Z}_t$;\\
Sample action $a_t$ from $p(a)$;\\
Calculate global reward $R_{global}$ by Eq.\ref{eq:R_global};\\
Calculate $\pi_{\theta}(a_t | o_t)$ and $V_{\pi}$;\\
Store $(o_t,\mathcal{Z}_t,a_t, \pi_{\theta}, V_{\pi})$ into $\mathcal{D}$ ;\\

\If{buffer length = Maximum buffer length $L_{max}$}
{
Sample a random minibatch of $S$ samples from $\mathcal{D}$; \\
Update the policy network with PPO-Clip;\\
}
}
}
  
\end{algorithm}

\section{Simulation and Performance Evaluation}
\label{section:6}
\subsection{Simulation Environment}
\label{simulation_env}

Our simulation platform is built based on an OpenAI Gym environment\cite{highway-env}. 
In the simulator, the actions determined by specific policies are translated to low-level steering and acceleration signals through a closed-loop PID controller. The longitude and lateral decisions of HVs are controlled by the IDM\cite{kesting2010enhanced} and MOBIL\cite{kesting2007general} models, respectively.
All HVs in our simulator are set with the constant-speed motion prediction and collision avoidance functions of the future $T_p s$.

\subsection{Simulation Settings}
In the DPL model, the encoder and decoder all contain one embedding layer and two GRU layers. The size of the embedding layer is 128,  and the hidden state size of the GRU is 256.
The steps of the trajectory we send to the DPL model per time is 20. We use a learning rate of $5 \times 10^{-4}$, and a 1024 batch size for training. The training epoch is set as 500.

As for the attention-based policy network, the encoder and decoder are MLP, which both have two linear layers and the size is $64 \times 64$. The size of the attention layer is 128, and the number of heads is 2. Meanwhile, the Deep Q-learning (DQN), Advantage Actor-Critic (A2C), and PPO algorithms are used as baselines in our experiments. The training parameters of RL algorithms are shown in Table.~\ref{tab:training_hyperparameter}.
Meanwhile, in order to observe the influence of different Coordination Tendencies on AV actions and system benefits, we sample every $\frac{\pi}{12}$ from 0 to $\frac{\pi}{2}$ and conduct experiments.

All simulation experiments are conducted in a computation platform with Intel Xeon Silver 4214R CPU, NVIDIA GeForce RTX 3090 GPU, and 128G Memory.

\begin{table}[!htbp]
    \centering
    \caption{The hyperparameter of the PPO Algorithm.}
    \label{tab:training_hyperparameter}
    \begin{tabular}{c c c}
        \toprule
        Symbol & Definition & Value\\
        \midrule
        $N_t$ & Total Training Steps & $10^5$ \\
        $S_u$ & Number of Forward Steps & 30 \\
        $\epsilon$ & PPO Clip Parameter & 0.2 \\
        $\lambda$ & Learning Rate & $10^{-4}$ \\
        $\gamma$ & Discount factor & 0.95\\
        $\tau$ & Target update rate	& 0.01\\
        $c_v$ & Value Loss Coefficient & 0.5 \\
        $c_e$ & Entropy Term Coefficient  &  0.01 \\
        $w_c$ & Weight for $r_c$ & 1 \\
        $w_e$ & Weight for $r_e$ & 1 \\
        $w_a$ & Weight for $r_a$ & 1 \\
        \bottomrule
    \end{tabular}
\end{table}

\subsection{Performance Evaluation}
\subsubsection{DPL Model}

\begin{figure}[!htbp]
    \centering
    \includegraphics[width=0.5\textwidth]{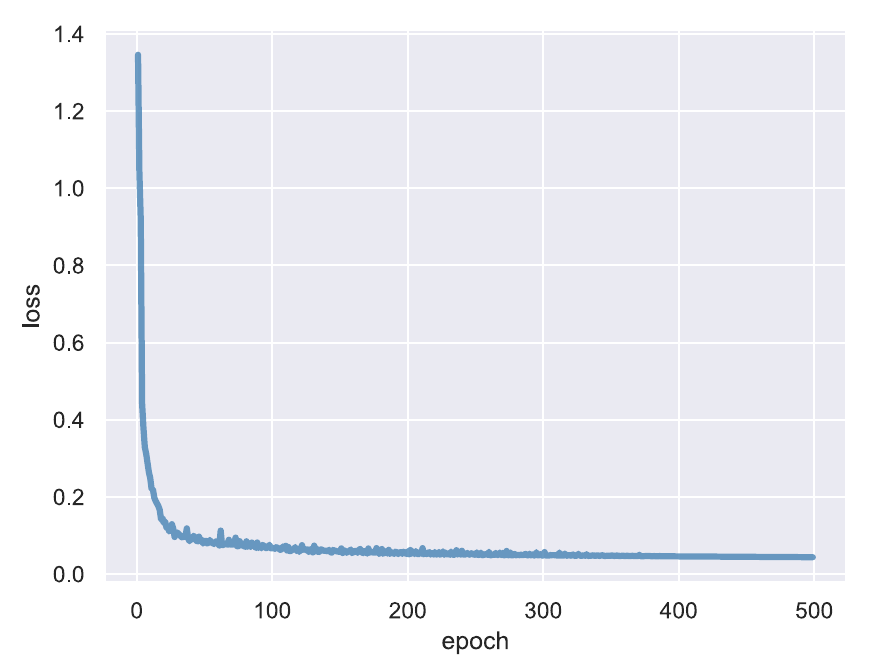}
    \caption{The training loss of Driving Prior Learning Model.
    }
    \label{fig:DPL_loss}
\end{figure}

The trajectory of the loss function for the DPL model is presented in Fig. \ref{fig:DPL_loss}. It is evident that the model achieves convergence at approximately 100 epochs. To ascertain the efficacy of the DPL model, we incorporated the inferred driving prior information into the training of the RL model and juxtaposed the outcomes against training without the integration of the DPL model. The trajectory of the average reward is illustrated in Fig. \ref{fig:train_reward}(a). Notably, the average reward of the AV demonstrates a significant increase, from $24.79$ to $28.13$, upon the assimilation of driving prior information. This represents a $13.47\%$ augmentation in the average reward, affirming the efficacy of the DPL model in enhancing AV performance.

\subsubsection{Prior Attention-PPO (PA-PPO) Model}
\begin{figure*}[!htbp]
    \centering
    \includegraphics[width=0.9\textwidth]{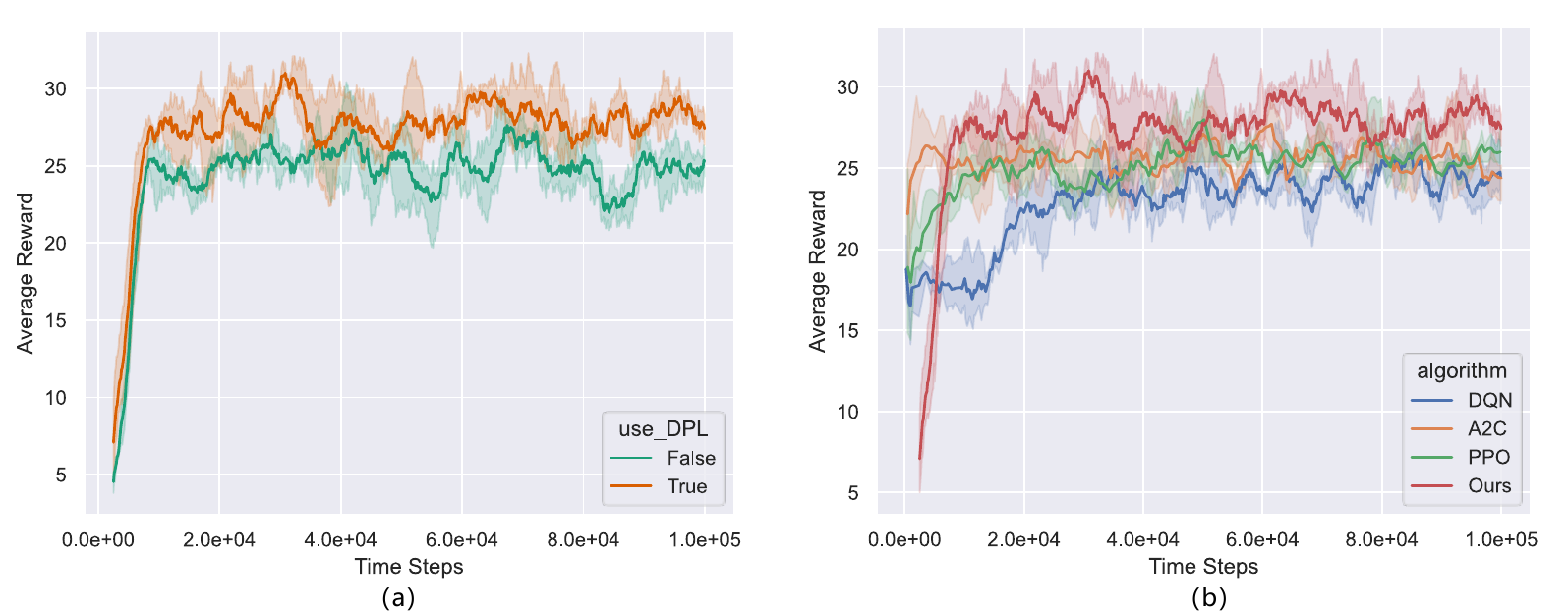}
    \caption{The average reward of the different algorithms, (a) whether to use the DPL model, (b) our algorithm and other baselines}
    \label{fig:train_reward}
\end{figure*}

The curves delineating the average rewards during the training of our PA-PPO algorithm, in conjunction with other baseline algorithms, are depicted in Fig. \ref{fig:train_reward}(b). It is discernible that while the PA-PPO algorithm initially exhibits lower rewards during the exploration phase, this phenomenon can be attributed to the initial stage's challenges in grasping the nuances of driver prior information, potentially leading to interpretational conflicts.

However, the PA-PPO algorithm swiftly transitions to an improved driving strategy around the time step $10^{4}$, maintaining commendable performance subsequently. By contrast, the DQN algorithm displays the least efficient exploration and suboptimal performance. While PPO and A2C algorithms showcase faster convergence rates, their overall efficacy remains inferior to that of PA-PPO.

Post $5 \times 10^{4}$ time steps, all algorithms effectively converge. Notably, at this juncture, the average reward for PA-PPO stands at 28.13, while corresponding figures for DQN, A2C, and PPO are 24.15, 25.70, and 25.73 respectively. This substantiates the superior performance of our algorithm in comparison to baseline approaches.

\subsection{Coordination Tendency Analysis}
\begin{figure}[!htbp]
    \centering
    \includegraphics[width=0.5\textwidth]{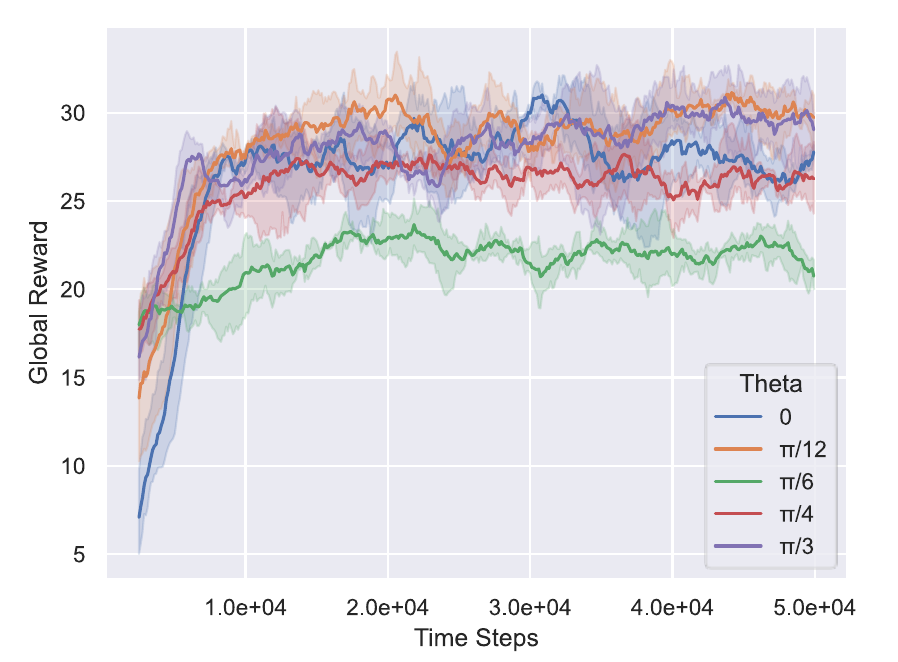}
    \caption{The comparison results of training average rewards with different $\phi$ .}
    \label{fig:CT_theta_compare}
\end{figure}

To elucidate the impact of varied CTs on training the AV's decision policy, we have conducted an analysis, with the average reward curves plotted in Fig. \ref{fig:CT_theta_compare}.

The findings illuminate the divergent effects engendered by distinct coordination tendencies. Utilizing the AV decision-making strategy devoid of Social Coordination Awareness as the baseline ($\phi=0$), the corresponding average reward registers at 26.49. When a subtle Coordination Tendency is introduced ($\phi=\frac{\pi}{12}$), commensurate improvements are witnessed in system benefits, as indicated by the elevated average reward of 29.44. This can be attributed to the modest incorporation of cooperative behaviors by AVs, which in turn curtails safety incidents like collisions, thus bolstering overall gains.

However, as the Coordination Tendency progressively intensifies ($\phi=\frac{\pi}{6}$), the linkage between AV cooperative conduct and an enhanced reward isn't linear or consistently positive; instead, it leads to a discernible reduction (average reward of 22.29). At this juncture, AV adopts more conservative strategies, prioritizing the welfare of other traffic participants, thereby mitigating systemic safety risks. Yet, the pronounced decline in AV's individual reward outweighs the surge in system reward, resulting in an overall reduction in reward magnitude.
\begin{figure}[!htbp]
    \centering
    \includegraphics[width=0.5\textwidth]{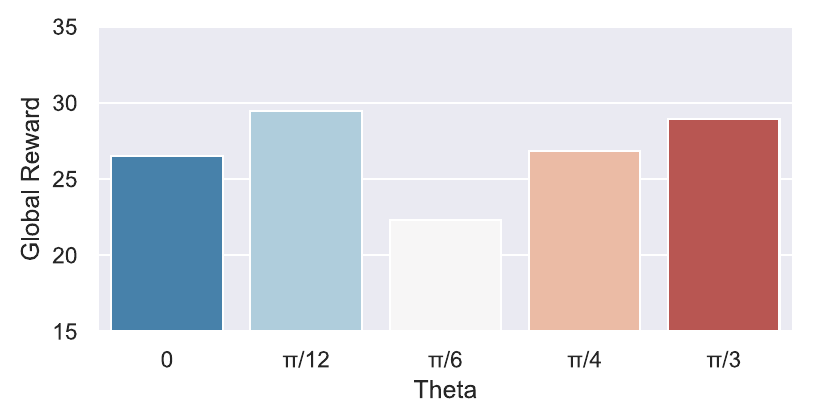}
    \caption{The mean training rewards of last $5 \times 10^4$ time steps with different $\theta$.}
    \label{fig:theta_ave_50000}
\end{figure}

As the coordination tendency further escalates ($\phi=\frac{\pi}{4}$, $\phi=\frac{\pi}{3}$), an uptick in the system's overall gains becomes evident (average rewards of 26.81 and 28.96 respectively). This upswing is attributable to the emergence of heightened altruistic behaviors exhibited by AVs, which culminate in enhanced benefits for fellow traffic participants. Moreover, the progressive augmentation of $\phi$ accentuates the weighting attributed to system benefits, thereby amplifying the collective advantage.

Simultaneously, we observe that an excessive coordination tendency isn't always advantageous. When $\phi>\frac{\pi}{3}$ (e.g., $\phi = \frac{5\pi}{12}$ or $\frac{\pi}{2}$), AV consistently opts for stationary actions to optimize its rewards, an approach that is manifestly untenable for AV's primary objectives. Consequently, convergence of the training process cannot be achieved in such cases, underscoring the necessity for judicious coordination tendency selection.

\subsection{Case Analysis}

\begin{figure*}[!htbp]
    \centering
    \includegraphics[width=1\textwidth]{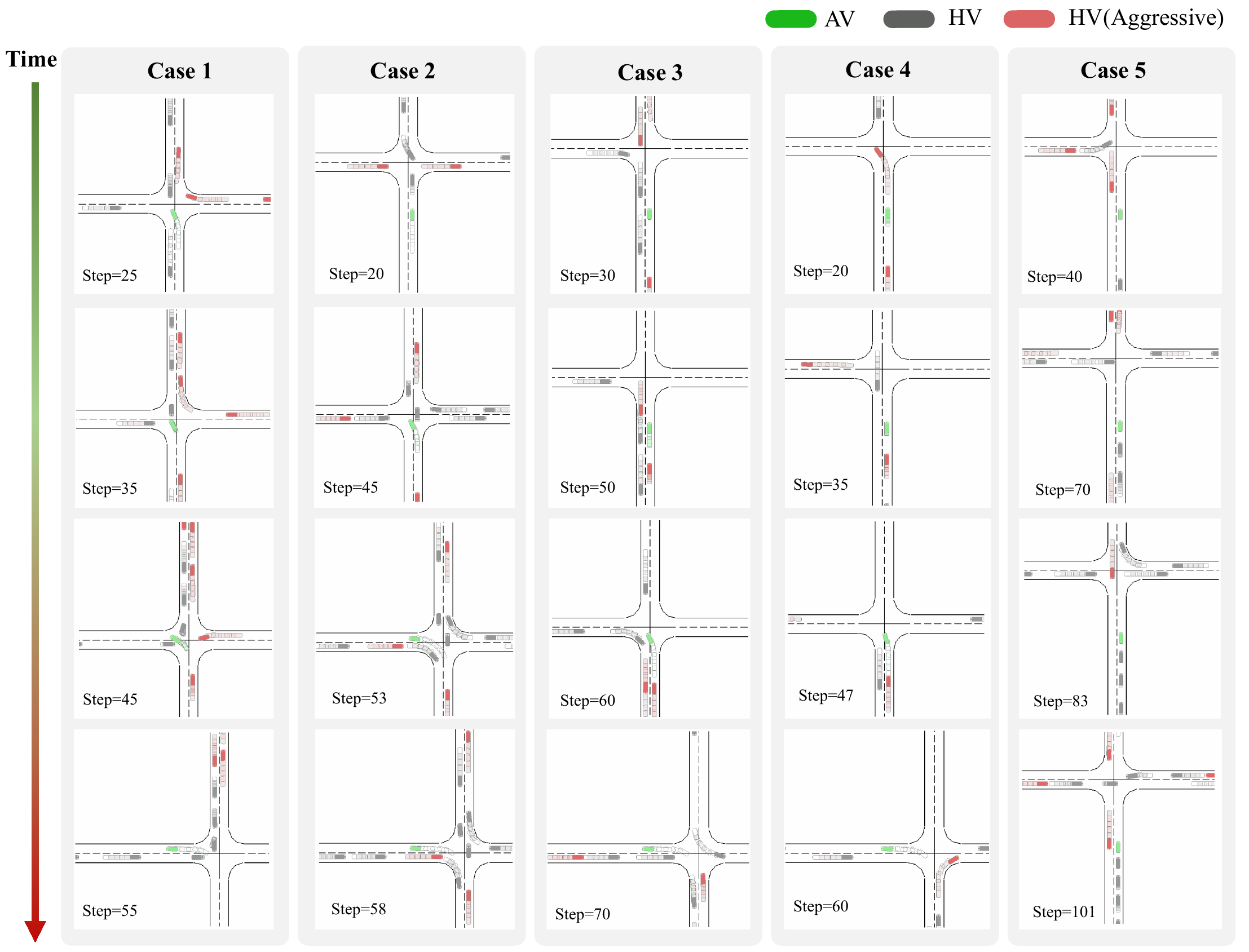}
    \caption{The snapshots of five interaction cases from different parameter settings, Case 1: CT $=0$ (AV competes with two HVs and rushes); Case 2: CT $=\frac{\pi}{12}$ (AV rushes to the conservative HV); Case 3: CT $=\frac{\pi}{12}$ (AV gives way to the aggressive HV); Case 4: CT $=\frac{\pi}{3}$ (AV yields conservatively to HV); Case 5: CT $=\frac{5\pi}{12}$ (Stopping still leads to congestion in the lane).}
    \label{fig:case_analysis}
\end{figure*}

\begin{figure}
    \centering
    \includegraphics[width=0.5\textwidth]{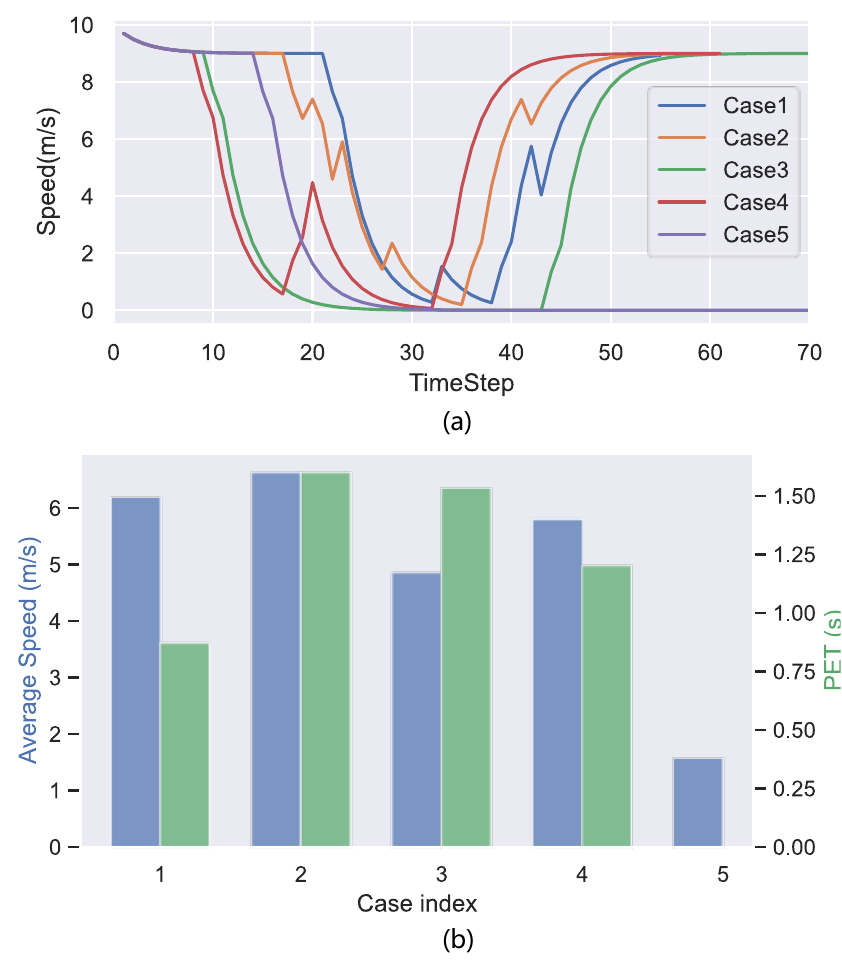}
    \caption{The interaction information of AVs from different cases: (a) The speed curve of AVs during the interaction, (b) The average speed and post encrochment time (PET) of AVs at the intersection.}
    \label{fig:case_velocity}
\end{figure}

Distinct coordination tendencies can yield varying action strategies for AVs. We have chosen five illustrative cases for analysis, detailed in Fig.\ref{fig:case_analysis} and Fig.\ref{fig:case_velocity}. 
The selected cases provide insights into AV strategies when encountering diverse human driver styles, with aggressive human drivers designated by the color red for enhanced visibility. 
The demo videos of these cases can be accessed at the site.\footnote{See \url{https://drive.google.com/drive/folders/1UW2UJv_ZpwLYPNvAV2uj6_j7MXdKfIVX?usp=sharing}}.

Case 1 highlights AV's behavior with a Coordination Tendency of $0$, signifying a self-centered orientation. Notably, at $\text{Time step}=25$, the AV navigates into the intersection and crosses paths with vehicles from multiple directions. During $\text{Time step}=35-45$, the AV decelerates, halts, and eventually maneuvers through the congested passage, effectively prioritizing its own interests. Although this strategy ensures intersection traversal, the resultant risky behavior runs counter to our desired safety standards, accentuating systemic safety risks.

Cases 2 and 3 introduce a subtle Coordination Tendency ($\text{CT}=\frac{\pi}{12}$). These cases showcase AV's adaptability in response to distinct human driver styles. In Case 2, AV encounters a conservative HV at $\text{Time step}=45$, promptly navigating through the intersection at $\text{Time step}=53$. In contrast, in Case 3, faced with an aggressive HV at $\text{Time step}=30$, AV chooses a judicious deceleration strategy ($\text{Time step}=30-50$), deftly avoiding conflict before proceeding ($\text{Time step}=60$). By harmonizing safety and efficiency, AV optimally navigates these scenarios, significantly enhancing intersection safety.

Case 4 raises the Coordination Tendency to $\frac{\pi}{3}$, prompting AV to overly prioritize other HV benefits. Even in a sparse traffic scenario (Case 4), and confronted by a cautious HV, AV exhibits unwarranted deference ($\text{Time step}=20-47$). This extended waiting period undermines efficiency and underscores the pitfalls of excessive consideration for other vehicles.

Elevating the Coordination Tendency further (Case 5, $\text{CT} = \frac{5\pi}{12}$), AV explores a distinct parking and waiting strategy ($\text{Time step}=40-101$) to maximize global benefits. The AV accords precedence to traffic from the other three lanes, prolonging its stay at the intersection, thereby disruputing the traffic flow significantly. This underscores the importance of circumspectly defining the Coordination Tendency within a judicious spectrum, thus averting undue traffic disturbances.

\section{Conclusion}
\label{section:7}
In the intricate landscape of human-machine mixed driving, the challenges of cultivating seamless human interaction and orchestrating efficient and accurate decisions persistently confront AVs. This study proposes a novel RL framework, interweaving driving priors and SCA  to elevate AV performance. Through a harmonious fusion of the DPL model, a policy network underpinned by multi-head attention mechanisms, and the SCA mechanism, our proposed framework empowers AVs to transcend conventional boundaries. This enables them to gain deeper insights into human driving tendencies, catalyze enhanced decision-making acumen, and engender behaviors characterized by heightened pro-social orientation.
The empirical evidence validates the efficacy of our framework, as it consistently outperforms baseline algorithms in terms of both reward acquisition and the manifest demonstration of socially coordinated behaviors.

Our future work will be dedicated to the augmentation of the learning and reasoning prowess of the DPL model, while further unraveling the latent coordination potential of AVs within traffic systems, all grounded on the bedrock of SCA principles. Furthermore, an extension of the RL framework's scope will be envisaged, endeavoring to surmount the challenges intrinsic to collaborative decision-making among multiple CAVs, alongside tackling complexities posed by diverse traffic scenarios. 

\bibliographystyle{IEEEtran} 
\bibliography{reference}

\end{document}